\documentclass[10pt,twocolumn,letterpaper]{article}

 \usepackage[pagenumbers]{cvpr} 

\usepackage{graphicx}
\usepackage{amssymb}

\usepackage{amsmath,epsfig}
\usepackage[table,xcdraw,dvipsnames]{xcolor}

\usepackage{float}
\usepackage{subcaption}
\usepackage{booktabs}
\usepackage{multirow}
\usepackage{enumitem}
\usepackage{pifont}
\usepackage{fancyhdr}
\usepackage[accsupp]{axessibility}

\makeatletter
\robustify\@latex@warning@no@line
\makeatother
\usepackage{authblk}
\usepackage[pagebackref,breaklinks,colorlinks]{hyperref}

\usepackage[capitalize]{cleveref}
\crefname{section}{Sec.}{Secs.}
\Crefname{section}{Section}{Sections}
\Crefname{table}{Table}{Tables}
\crefname{table}{Tab.}{Tabs.}

\def\cvprPaperID{35} 
\def\confName{CVPR}
\def\confYear{2023}

\begin{document}

\title{Diversity is Definitely Needed: Improving Model-Agnostic Zero-shot Classification via Stable Diffusion}
\author[1]{Jordan Shipard}
\author[1,2]{Arnold Wiliem}
\author[1]{Kien Nguyen Thanh}
\author[3]{Wei Xiang}
\author[1]{Clinton Fookes}

\affil[1]{\small Signal Processing, Artificial Intelligence and Vision Technologies (SAIVT), Queensland University of Technology, Australia}
\affil[2]{Sentient Vision Systems, Australia}
\affil[3]{School of Computing, Engineering and Mathematical Sciences, La Trobe University, Australia}
\affil[ ]{\textit {\tt \small \{jordan.shipard@hdr., k.nguyenthanh@, c.fookes@\}qut.edu.au}, \textit{ \tt \small arnoldw@sentientvision.com}, \textit{ \tt \small W.Xiang@latrobe.edu.au}}


\maketitle

\begin{abstract}

In this work, we investigate the problem of Model-Agnostic Zero-Shot Classification (MA-ZSC), which refers to training non-specific classification architectures (downstream models) to classify real images without using any real images during training. Recent research has demonstrated that generating synthetic training images using diffusion models provides a potential solution to address MA-ZSC. However, the performance of this approach currently falls short of that achieved by large-scale vision-language models. One possible explanation is a potential significant domain gap between synthetic and real images. Our work offers a fresh perspective on the problem by providing initial insights that MA-ZSC performance can be improved by improving the diversity of images in the generated dataset. We propose a set of modifications to the text-to-image generation process using a pre-trained diffusion model to enhance diversity, which we refer to as our \textbf{bag of tricks}. Our approach shows notable improvements in various classification architectures, with results comparable to state-of-the-art models such as CLIP. To validate our approach, we conduct experiments on  CIFAR10, CIFAR100, and EuroSAT, which is particularly difficult for zero-shot classification due to its satellite image domain. We evaluate our approach with five classification architectures, including ResNet and ViT. Our findings provide initial insights into the problem of MA-ZSC using diffusion models. All code is available at \url{https://github.com/Jordan-HS/Diversity_is_Definitely_Needed}
\end{abstract}

\vspace{-5mm}
\section{Introduction}
\begin{figure}
    \centering
    \includegraphics[clip, trim=6.8cm 11.7cm 3.7cm 5.7cm, scale=0.7]{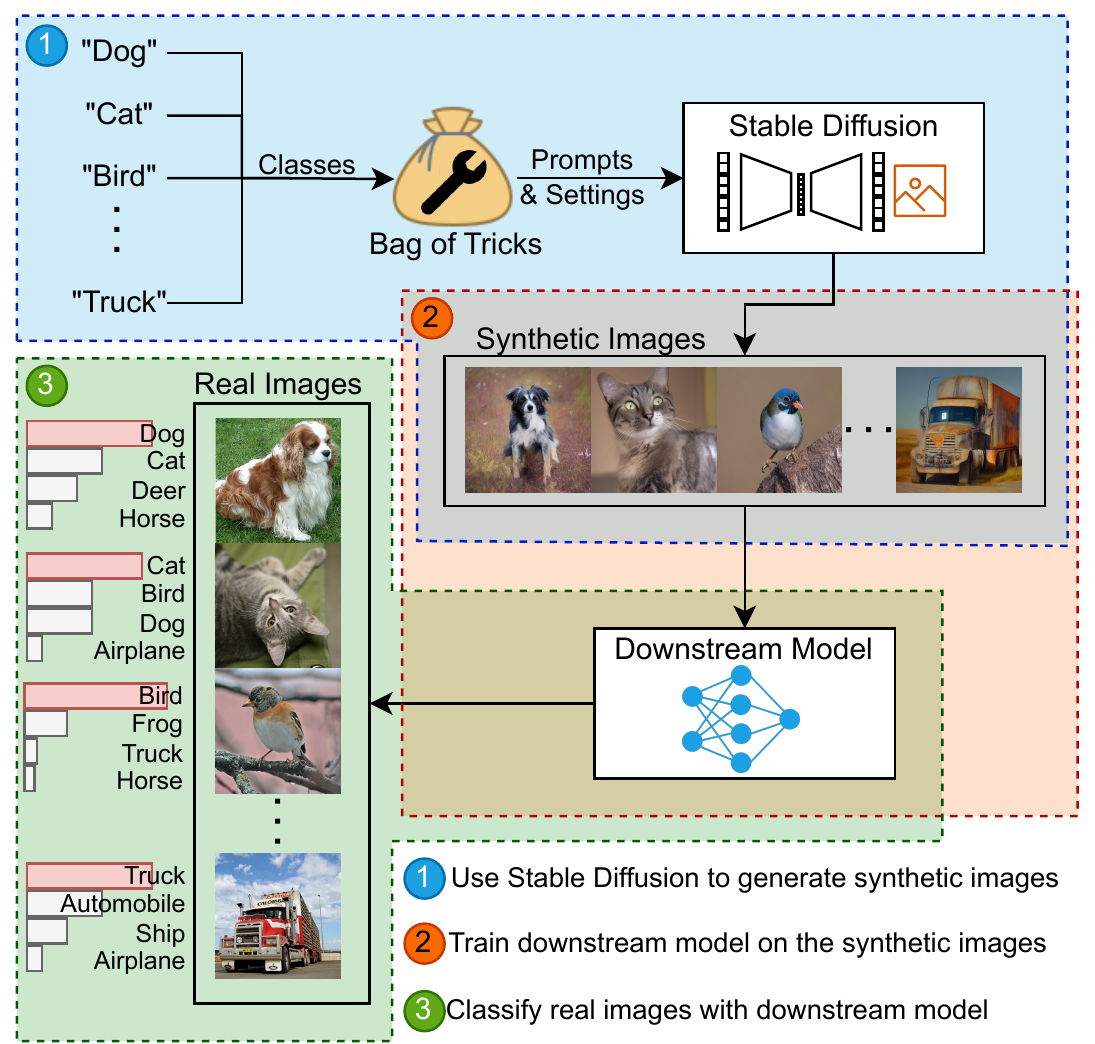}
    \caption{A diagram of our proposed model agnostic zero-shot classification method. We first generate a set of diverse training images using our proposed \textit{bag of tricks}. These generated synthetic images are then used to train a downstream model capable of classifying a set of real images. This method achieves zero-shot performance comparable to CLIP~\cite{radford_learning_2021}.}
    \label{fig:diagram}
\end{figure}
Data is a critical element for the successful training of deep learning models \cite{radford_language_nodate,brown_language_2020}. However, acquiring and curating a high-quality dataset can be challenging, time-consuming, and expensive \cite{schuhmann_laion-5b_nodate, deng_imagenet_2009}. This is especially true for inherently expensive domains, such as remote sensing where the creation of a dataset requires the use of satellites \cite{sumbul_bigearthnet_2019}. The zero-shot learning tackles the data issue by training a model using no data from the downstream task~\cite{wang_survey_2019}. Generally, zero-shot learning models use specialised architecture~\cite{radford_learning_2021, novackCHiLSZeroShotImage2023, udandaraoSuSXTrainingFreeNameOnly2023, menonVisualClassificationDescription2022, prattWhatDoesPlatypus2022} or learn auxiliary features to achieve the objective \cite{patterson_sun_2012}. This limits the array of potential model architectures. On the other hand, the Model-Agnostic Zero-Shot Learning problem aims to address this issue by considering methods that allow any model architecture to perform the zero-shot learning task. In this work, we specifically investigate image classification tasks, and as such, refer to the problem domain as Model-Agnostic Zero-Shot Classification (MA-ZSC). Solving the MA-ZSC problem alleviates the previously mentioned problems of constructing a high-quality dataset without the need for specialised architecture or techniques applied to the model.

One avenue to address the MA-ZSC problem is by generating a synthetic dataset which is then used to train a classification model. A Recent work \cite{he_is_2022} explored the use of generated synthetic images powered by the GLIDE diffusion model~\cite{nichol_glide_2022} primarily for improving the zero-shot and few-shot performance of CLIP~\cite{radford_learning_2021}; however, they briefly explore training classification models, finding synthetic images to be inferior compared to real images. This is thought to be due to a domain gap between real and synthetic images \cite{he_is_2022}. We revisit the issue and find it possible to improve the MA-ZSC performance by increasing the diversity of the images in the generated datasets. Increasing diversity in training data has been utilised in various domains, such as computer vision domains \cite{yueDomainRandomizationPyramid2019} and robotics \cite{tobinDomainRandomizationTransferring2017}. For instance, in domain randomisation methods in the robotics domain \cite{tobinDomainRandomizationTransferring2017}, increasing diversity in the training data would help to reduce the 'reality gap' that separates simulated robotics from experiments on hardware. With enough variability in the simulator, the real world may appear to the model as just another variation. Increasing training data variability is also a common practice in training classification models by employing data augmentation methods \cite{shortenSurveyImageData2019}. The variability of the training data is increased by simply performing data augmentation operations such as flipping, and rotations. Figure \ref{fig:diagram} presents our overall method, and we present our results as follows.

Using the latent diffusion model, Stable Diffusion \cite{noauthor_high-resolution_nodate}, we start by generating baseline synthetic datasets using the prompt ``an image of a \{class\}" for CIFAR10 \cite{krizhevsky_learning_2009}, CIFAR100 \cite{krizhevsky_learning_2009} and EuroSAT \cite{helber_Eurosat_2019}, which is particularly difficult for zero-shot classification due to its satellite image domain, as noted in \cite{radford_learning_2021}. For each dataset, \{class\} is replaced iteratively with the class labels to generate images belonging to that class. We then train a ResNet50 model \cite{he_deep_2015} on these datasets and obtain a zero-shot top-1 accuracy of \textbf{60.1\%}, \textbf{29.72\%} and \textbf{36.18\%} respectively. For reference, the CLIP zero-shot performance with a ResNet50 backbone achieves 75.6\%, 41.6\%, 41.1\%.

In order to improve the diversity of images in the generated dataset we propose a set of modifications to the text-to-image generation process, which we refer to as our \textit{bag of tricks}. Each trick only improves image diversity, with no mitigation of the image domain. In fact, one of our tricks (\textit{multi-domain}) specifically generates out-of-domain examples. After applying the proposed  \textit{bag of tricks} and finding the best tricks for each dataset, we obtain top-1 zero-shot classification accuracy of \textbf{81\% (\textcolor{Green}{\(\uparrow\)20.5}}) on CIFAR10,  \textbf{45.63\% (\textcolor{Green}{\(\uparrow\)15.91})} on CIFAR100 and \textbf{42.5\% (\textcolor{Green}{\(\uparrow\)6.41})} on EuroSAT. Surprisingly, these results surpass the performance of CLIP-ResNet50~\cite{radford_learning_2021}. 

\noindent
We list our \textbf{contributions} as follows: 
\begin{enumerate}[noitemsep, topsep=0pt]
    \item Equipped with insights found in our work, we demonstrate the feasibility of addressing the MA-ZSC problem by training classification models on a high-quality Stable Diffusion~\cite{noauthor_high-resolution_nodate} generated synthetic dataset.
    \item We show that improving the diversity of images in a generated synthetic dataset improves the MA-ZSC performance. 
    \item We provide a \textit{bag of tricks} for improving diversity during latent diffusion image generation. 
\end{enumerate}

\noindent We continue our paper as follows. In Section \ref{sec:related_works}, we provide a summary of previous works related to zero-shot learning, image generation, and training with synthetic data. We then describe the problem of model-agnostic zero-shot classification, and introduce our solution in Section~\ref{sec:MAZSC}, before describing the process of how latent diffusion models generate images in Section \ref{sec:gen-imgs}. In Section \ref{sec:improve}, We investigate how we can improve the quality of our generated synthetic datasets. Following this, we propose our \textit{bag of tricks} to improve the diversity of our generated synthetic datasets in Section \ref{sec:bag_of_tricks}. Finally, we present our experimental results on CIFAR10~\cite{krizhevsky_learning_2009}, CIFAR100\cite{krizhevsky_learning_2009}, and EuroSAT~\cite{helber_Eurosat_2019} in Section \ref{sec:experiments}, demonstrating the impact of our proposed \textit{bag of tricks} on zero-shot performance across five classification models.

\section{Related Work}
\label{sec:related_works}
In this section, we first show previous works relating to zero-shot learning and the introduction of CLIP; before covering the recent history of image generation and diffusion models. Lastly, we will cover previous works where training was done using generated synthetic images.

\subsection{Zero-shot Learning and CLIP}
Zero-shot learning (ZSL) refers to the ability of a trained model to classify classes it was not trained on \cite{wang_survey_2019}. For image classification, this is traditionally achieved by learning auxiliary attributes instead of pre-defined classes \cite{patterson_sun_2012,xian_zero-shot_2017}. However, a more recent and effective approach is to train an image and text encoder to learn joint representations, such as with CLIP~\cite{radford_learning_2021}.
By encoding an image and names or descriptors of the target datasets classes, zero-shot predictions can be made by finding which descriptions text embedding is the most similar to the image embedding. Using this CLIP achieves zero-shot performance on datasets such as ImageNet \cite{deng_imagenet_2009} (76.2\%) comparable to supervised training of high-quality models \cite{he_deep_2015, dosovitskiy_image_2021}. Several works look to improve the ZSL performance of CLIP by improving the text description used \cite{novackCHiLSZeroShotImage2023, udandaraoSuSXTrainingFreeNameOnly2023, menonVisualClassificationDescription2022, prattWhatDoesPlatypus2022}. In this work, we focus on improving the zero-shot learning performance of non-specific architectures, specifically focusing on image classification. We refer to this problem as Model Agnostic Zero-Shot Classification (MA-ZSC).

\subsection{Image Generation and Diffusion Models}
Text-to-image generative models saw a significant performance improvement with DALL-E 1 \cite{ramesh_zero-shot_2021}. The next generation of generative models, GLIDE \cite{nichol_glide_2022}, Latent Diffusion Models (LDM) \cite{ramesh_hierarchical_2022}, DALL-E 2 \cite{noauthor_high-resolution_nodate}, and Imagen \cite{saharia_photorealistic_nodate}, use text encoders paired with diffusion models, in contrast to the discrete variational autoencoder and autoregressive transformer used by DALL-E 1. Diffusion models produce an image by denoising Gaussian noise according to some provided conditioning, such as a text prompt. Latent diffusion models perform the denoising in the latent space, making them more computationally efficient. This work uses a latent diffusion model to generate synthetic images.

\subsection{Training with Synthetic Images}
Since our research's inception, several papers have investigated utilizing diffusion models to extend real datasets for domain adaptation \cite{NotJustPretty2022}, semi-supervised learning~\cite{DiffusionModelsSemiSupervised2023}, or generating data augmentations~\cite{EffectiveDataAugmentation2023a, LeavingRealityImagination2023}. In our work, we concentrate solely on training downstream models on entirely synthetic generated datasets. This differs from the two closest works to ours \cite{he_is_2022, besnier_this_2020}, as He et al.~\cite{he_is_2022} studies if synthetic images from a diffusion model, GLIDE \cite{nichol_glide_2022}, can be used to fine-tune CLIPs zero-shot and few-shot performance. With respect to training downstream models on synthetic data they conclude synthetic images are \(5\times\) less data efficient than real images. Additionally, He et al. \cite{he_is_2022} show synthetic images can effectively pre-train a classifier, on par with ImageNet pre-training.  While Besnier et al.~\cite{besnier_this_2020} use a GAN pre-trained on ImageNet \cite{deng_imagenet_2009} and propose strategies for improving the training quality of the generated images. With their improvements they were able to achieve 88.8\% accuracy on ImageNet-10 when training on synthetic images, compared to 88.4\% on real images. In our work, we generate training images using a diffusion model trained on a much larger dataset and distribution of images.

\begin{figure}[t]
\centering
    \begin{subfigure}[b]{0.11\textwidth}
        \includegraphics[scale=0.115]{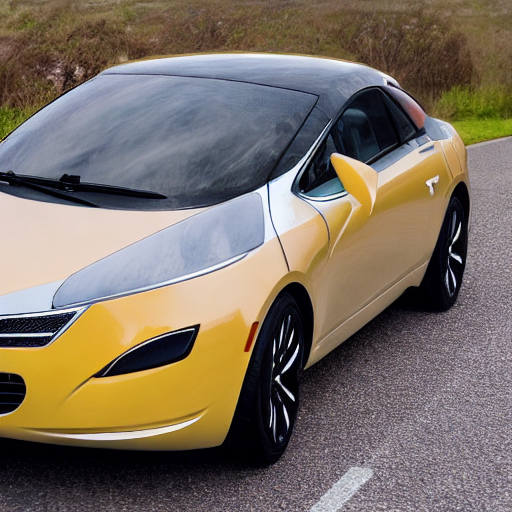}
        \includegraphics[scale=0.115]{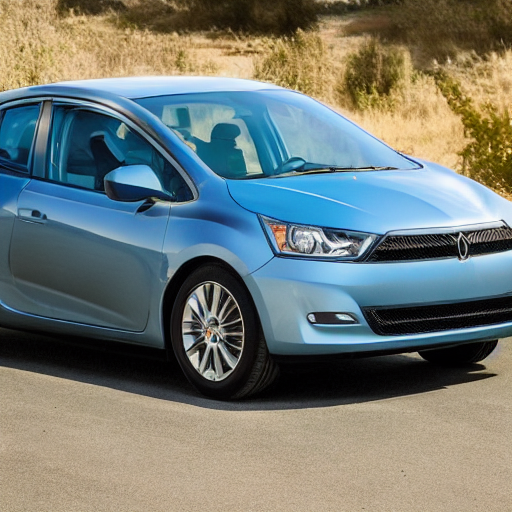}
        \caption{photo}
        \label{subfig:img}
    \end{subfigure}
    \hfill
    \begin{subfigure}[b]{0.11\textwidth}
        \includegraphics[scale=0.115]{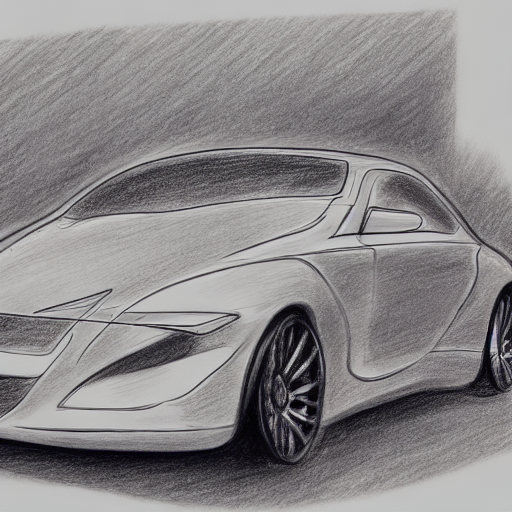}
        \includegraphics[scale=0.115]{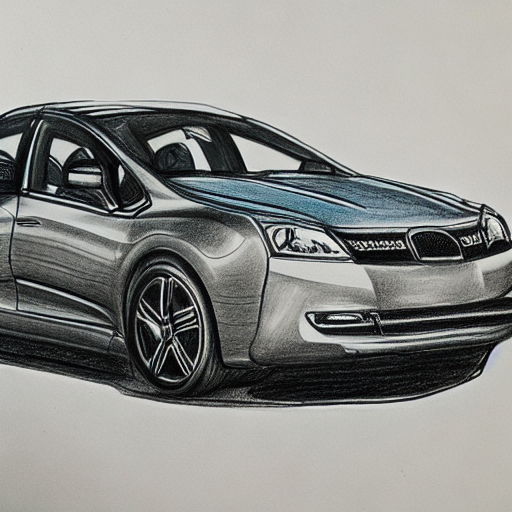}
        \caption{drawing}
        \label{subfig:drawing}
    \end{subfigure}
    \hfill
    \begin{subfigure}[b]{0.11\textwidth}
        \includegraphics[scale=0.115]{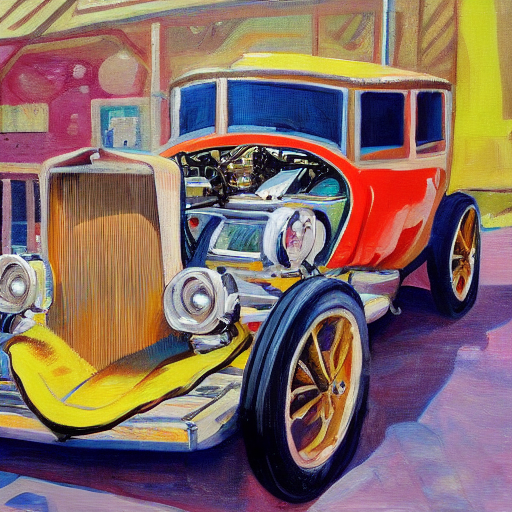}
        \includegraphics[scale=0.115]{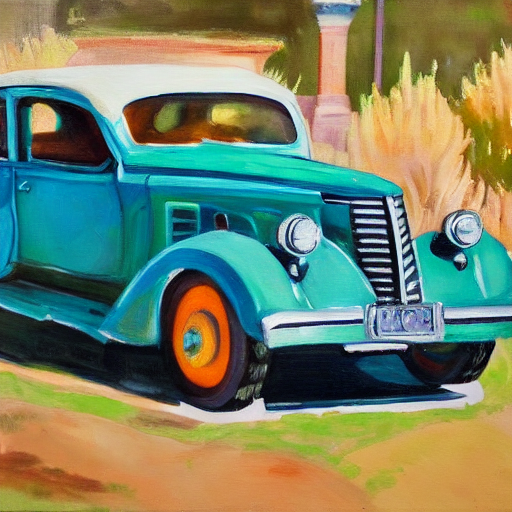}
        \caption{painting}
        \label{subfig:painting}
    \end{subfigure}
    \hfill
    \begin{subfigure}[b]{0.11\textwidth}
        \includegraphics[scale=0.115]{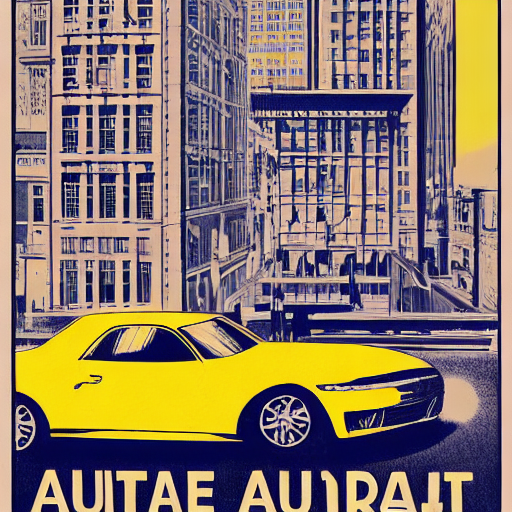}
        \includegraphics[scale=0.115]{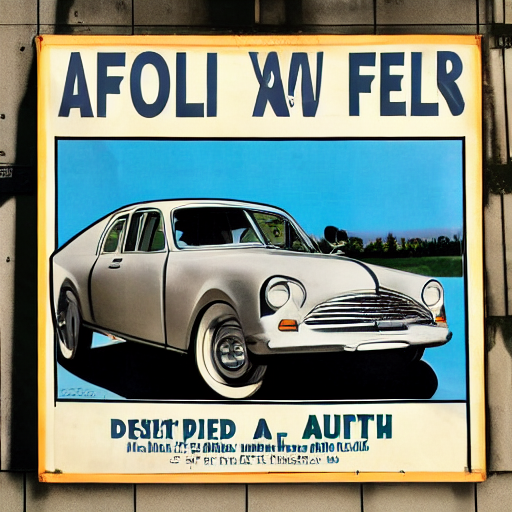}
        \caption{poster}
    \end{subfigure}
    \caption{Examples of images generated from the prompt ``a \{caption\} of a car", where \{caption\} is the caption of each sub-figure. Each row shares a common initial Gaussian.}
    \label{fig:txt2img}
\end{figure}

\section{Model-Agnostic Zero-Shot Classification}
\label{sec:MAZSC}
\textbf{Z}ero-\textbf{S}hot \textbf{L}earning (ZSL), as traditionally defined by \cite{wang_survey_2019}, aims to learn a classifier $f^{u}(\cdot):\mathcal{X}\rightarrow \mathcal{U}$, trained on labelled training instances $D^{tr}$ belonging a set of seen classes $\mathcal{S}$ (\(\mathcal{S}=\{c_i^s | i=1, ..., N_s\}\)), to classify testing instances $X^{te}$ belonging to a set of unseen classes $\mathcal{U}$ (\(\mathcal{U}=\{c_i^u | i=1, ..., N_u\}\)). Where $\mathcal{S}\cap\mathcal{U}=\varnothing$. 
One method to address this problem is to describe each unseen class $c_i^u$ with a set of image attributes extracted from the images in the seen classes $\mathcal{S}$~\cite{lampert_zero-shot_2014}.
To classify an unseen image, we first extract image attributes from the image. Then the classification is done in the attribute feature space by comparing 
the image attributes of the unseen image with each unseen class image attributes.
With the advent of vision-language models such as CLIP~\cite{radford_learning_2021}, the image attributes are replaced by the natural language text which is more expressive and can more accurately describe the unseen classes.
In the CLIP model, the text features and image features are correlated. Thus, we can classify an unseen image by measuring the dot 
product between its image features, and the text features extracted from each class description.

Whilst the CLIP model has shown impressive zero-shot performance, one still needs to use the CLIP model and its zero-shot methodology. 
We argue that this limits the applicability of the zero-shot classification. For instance, it is non-trivial to deploy the CLIP model into 
edge devices due to 
its large model size and complexity.
This motivates us to consider the \textbf{M}odel-\textbf{A}gnostic \textbf{Z}ero-\textbf{S}hot \textbf{C}lassification (MA-ZSC) problem.
In the MA-ZSC setting, we wish to use any non-specific architecture and methodology to perform the zero-shot classification task.
In other words, any classification model and methodology can be used for $f^u(\cdot)$. 

To address the MA-ZSC problem, we utilise a \textbf{L}atent \textbf{D}iffusion \textbf{M}odel~\cite{noauthor_high-resolution_nodate} (LDM) which can generate synthetic training images for each unseen class from its textual description. Once the synthetic training dataset, $D^{syn}$, are generated, we can then train any downstream classification model. 
Note that we are not claiming that our work is the first work to employ a synthetic image generation strategy to address the MA-ZSC problem. Rather, we show the feasibility of the strategy by combining the LDM with our proposed \textit{bag of tricks} to generate more diverse synthetic images.

\section{Generating Training Images}
\label{sec:gen-imgs}
In our work, we use a Latent Diffusion Model \cite{noauthor_high-resolution_nodate} loaded with Stable Diffusion V1.4 weights to generate synthetic images. Stable Diffusion was trained on a subset of the LAION-5B dataset \cite{schuhmann_laion-5b_nodate} and generates images of \(512\times512\) pixels, which we resize to the native image size of each dataset. Stable Diffusion uses a frozen CLIP ViT-L/14 text encoder to provide conditioning from text prompts, similar to \cite{noauthor_high-resolution_nodate, saharia_photorealistic_nodate}. Although LDMs can generate images from a number of different conditionings (image, text, semantic map), in our work we only generate images via text prompt. In the following sections, we will first describe text-to-image generation in more detail before investigating the potential domain gap between real and synthetic images. Finally, we hypothesise image diversity is more important for improving zero-shot classification and provide a \textit{bag of tricks} for improving the synthetic image diversity.

\subsection{Text-to-Image}
For text-to-image generation, a prompt that describes the desired contents of the image is required to guide the diffusion process. The prompt is projected to an intermediate representation via CLIP's text encoder and then mapped to the intermediate layers of the LDMs denoising UNet via cross-attention \cite{noauthor_high-resolution_nodate}. The LDM uses this guidance to diffuse a latent representation of an image starting from Gaussian noise. The resulting latent representation is then decoded back into the pixel domain to produce the final image. Figure \ref{fig:txt2img} shows examples of generated images using different text prompts.

There are three main hyperparameters that control the generation process. \textbf{DDIM Steps} controls the number of steps taken by the \textbf{D}enoising \textbf{D}iffusion \textbf{I}mplicit \textbf{M}odel \cite{song_denoising_2022} in the denoising process. More steps generally result in more realistic and coherent images, while fewer result in more disjointed surreal images. The images in Figure \ref{fig:txt2img}, and all synthetic images used in this work, were generated with 40 DDIM steps. \textbf{Unconditional Guidance Scale (UGC)} controls the scale between the precision of the generated image matching the provided prompt and generation diversity. This is done by scaling between the jointly trained conditional and unconditional diffusion models \cite{ho_classifier-free_2022}. A lower UGC value means less guidance and therefore more diversity and vice versa. Lastly, there is the \textbf{Seed} from which the initial Gaussian is generated, and serves as the starting point for diffusion. The images in each row of Figure \ref{fig:txt2img} were all generated from common seeds, resulting in the cars in each row sharing similar features, such as car shape, position and colour.

\subsection{Improving Synthetic Data for Training}
\label{sec:improve}
Before attempting to construct a high-quality synthetic dataset we first validate that our synthetic images contain enough semantic features for classification. In a prior work, He et al.~\cite{he_is_2022} conclude there is a domain gap between real and synthetic images generated via their chosen diffusion model GLIDE~\cite{nichol_glide_2022}. They suggest that reducing this gap is necessary to improve the quality of synthetic images for training purposes. If the domain gap is significant we expect it to noticeably impact our initial validation experiment. If it does not, however, we then need to investigate what factors do impact the quality of a synthetic dataset. 


\begin{figure}[t]
    \centering
    \includegraphics[scale=0.25]{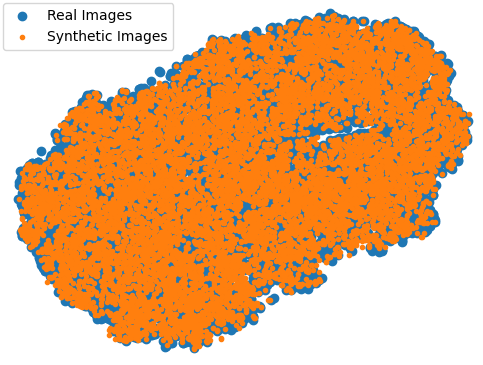}

    \caption{T-SNE plot comparing clustering of real and synthetic images. The lack of separation in the feature space suggests that the synthetic images contain semantically meaningful features.}
    \label{fig:synVsRealTSNE}
\end{figure}

\begin{figure}[t]
    \centering
    \hspace{-8pt}
    \begin{subfigure}[b]{0.2\textwidth}
        \includegraphics[scale=0.2]{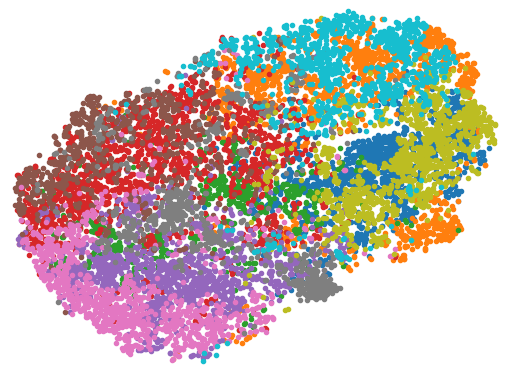}
        \caption{Real images.}
    \end{subfigure}
    \begin{subfigure}[b]{0.2\textwidth}
        \includegraphics[scale=0.2]{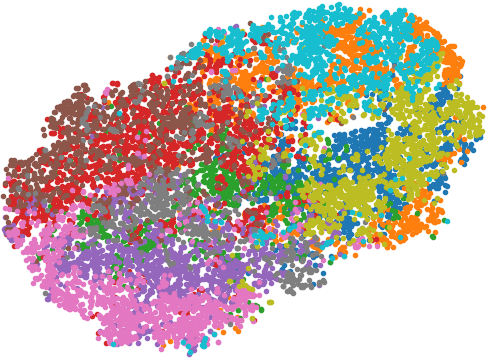}
        \caption{Synthetic images.}
    \end{subfigure}
    \hspace{-5pt}\includegraphics[scale=0.25]{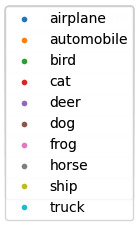}
    \caption{Two t-SNE plots showing the clustered classes for real and synthetic images. The relative classes are clustered together, demonstrating the ability of the classifier to classify synthetic images despite only being trained on real images.}
    \label{fig:ClassesTSNE}
\end{figure}

\subsubsection{Validating the Usefulness of Synthetic Images}
One approach for validating the quality of synthetic images is the Fréchet Inception Distance (FID) score \cite{GANsTrainedTwo2017}. Recent generative diffusion models achieve FID scores as low as 7.27 \cite{saharia_photorealistic_nodate} on the MS-COCO dataset \cite{lin2014microsoft}, with Stable Diffusion (V1.4) achieving a score of 16. For reference, real images overlayed with 25\% Gaussian noise result in FID scores of approximately 50 \cite{GANsTrainedTwo2017}. With this in mind, we conjecture that the low FID scores of recent diffusion models suggest they are all capable of generating realistic images. In order to examine if synthetic images contain semantically meaningful features we take a ResNet50 model \cite{he_deep_2015} with pre-trained ImageNet weights \cite{deng_imagenet_2009} and fine-tune only the classifier head on the real CIFAR10 \cite{he_learning_2009} dataset. We then visualise the feature space of the real images and our generated synthetic CIFAR10 images using a t-SNE plot \cite{van2008visualizing}. If the synthetic images contain vastly different semantic features compared to real images we expect the model to fail in classifying them. Additionally, we expect the real and synthetic images to occupy different areas of the feature space. However, in Figure \ref{fig:synVsRealTSNE} we can see the features of the real and synthetic images are intermingled. Furthermore, in Figure \ref{fig:ClassesTSNE} we see the classes are closely clustered, with the model correctly classifying \textbf{76.61\%} of the real images and \textbf{63.8\%} of the synthetic images. This initial experiment validates that the synthetic images contain enough meaningful semantic features and therefore should be useful as training images.

\begin{figure}[t]
    \centering
    \begin{subfigure}[c]{0.5\textwidth}
    \centering
        \includegraphics{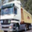}
        \includegraphics{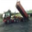}
        \includegraphics{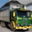}
        \caption{Original images}
        \label{subfig:orig_images}
    \end{subfigure}
    
    \begin{subfigure}[b]{0.5\textwidth}
    \centering
        \includegraphics[scale=0.21]{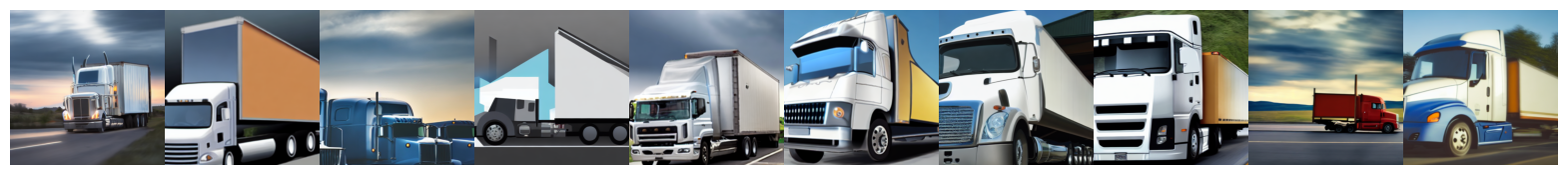}
        \includegraphics[scale=0.21]{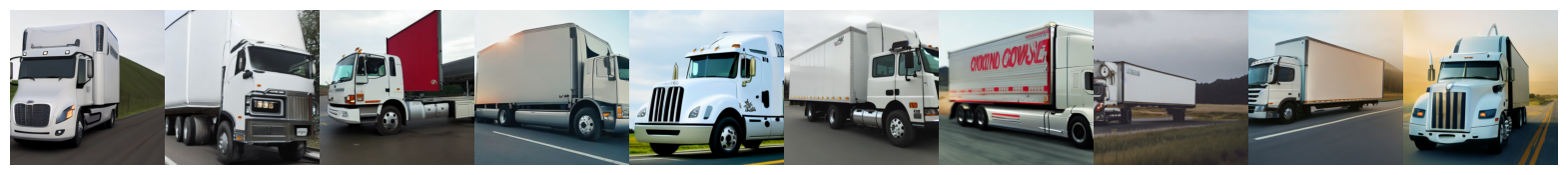}
        \includegraphics[scale=0.21]{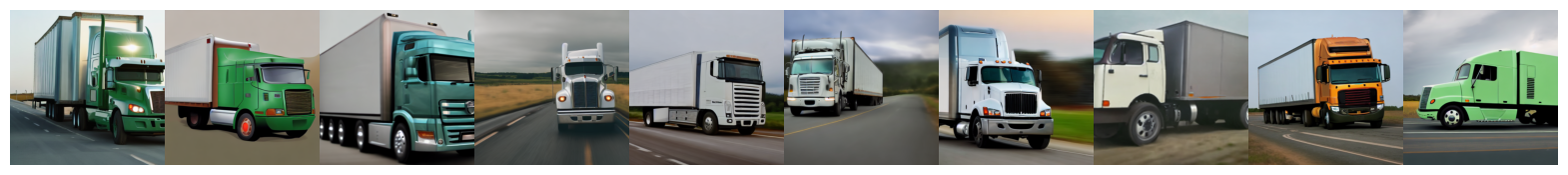}
        \includegraphics[scale=0.21]{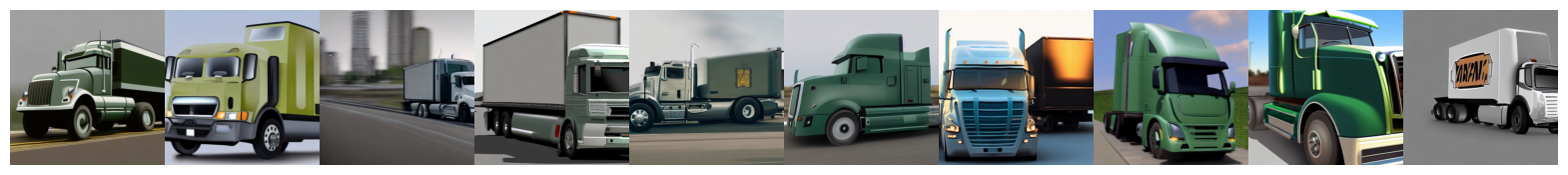}
        \caption{Synthetic images from latent interpolation}
        \label{subfig:syn_images}
    \end{subfigure}
    
    \caption{Examples of images generated from the combination of different initial latents. Row 1 of (b) is from the left most image of (a). Row 2 is the average of the centre and left most images. Row 3 is the average of all three images in (a). Row 4 is random combinations of all three images in (a) }
    \label{fig:latentDiversity}
\end{figure}

\begin{figure}[t]
    \centering
    \begin{subfigure}[b]{0.2\textwidth}
    \centering
        \includegraphics[scale=0.25]{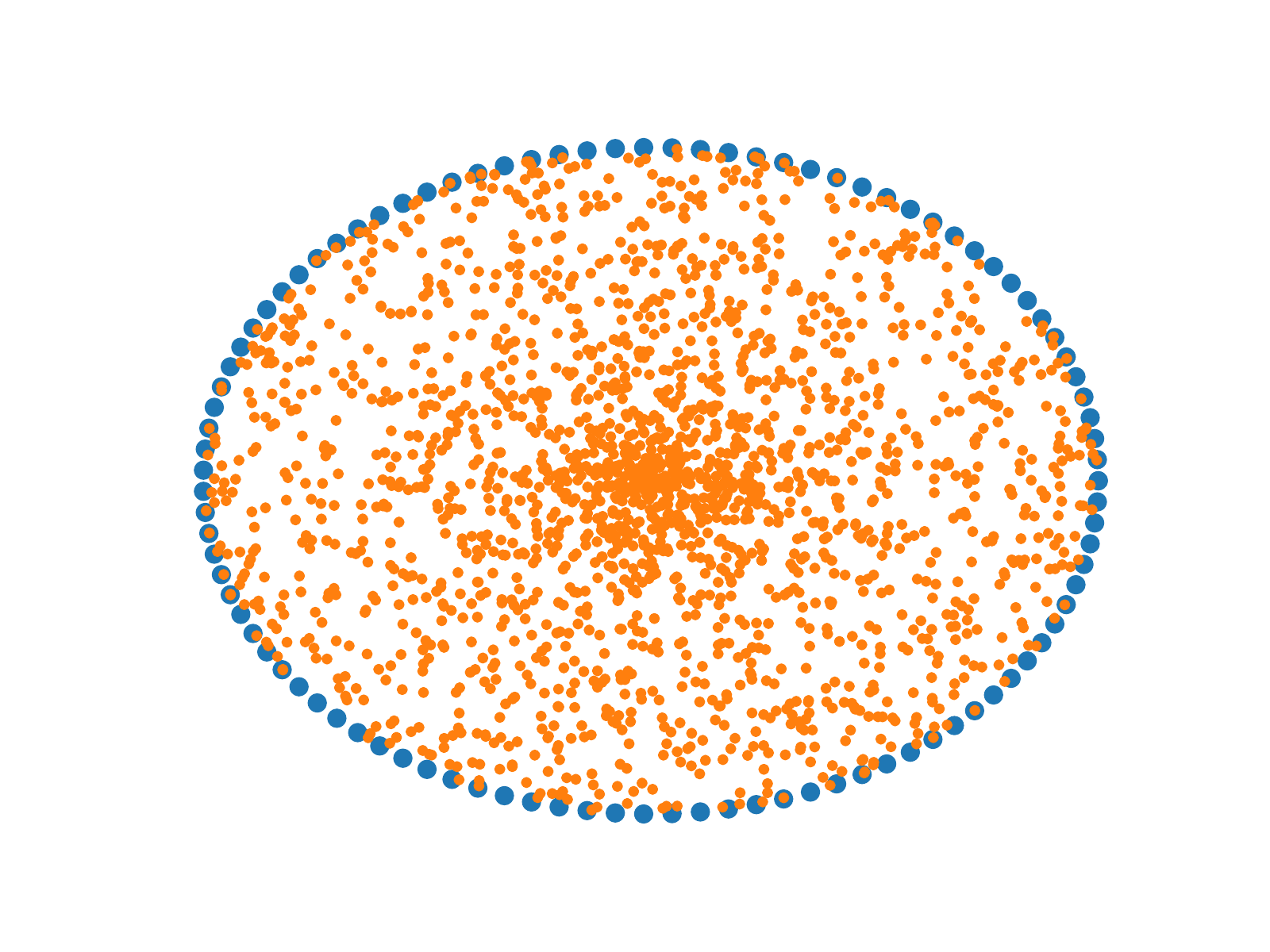}\vspace{-10pt}
        \includegraphics[scale=0.35]{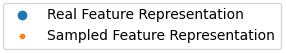}
        \caption{Sampled linearly interpolated feature representations from between all real feature representations.}
        \label{subfig:60_lat}
    \end{subfigure}
    \hspace{10pt}
    \begin{subfigure}[b]{0.2\textwidth}
    \centering
        \includegraphics[scale=0.25]{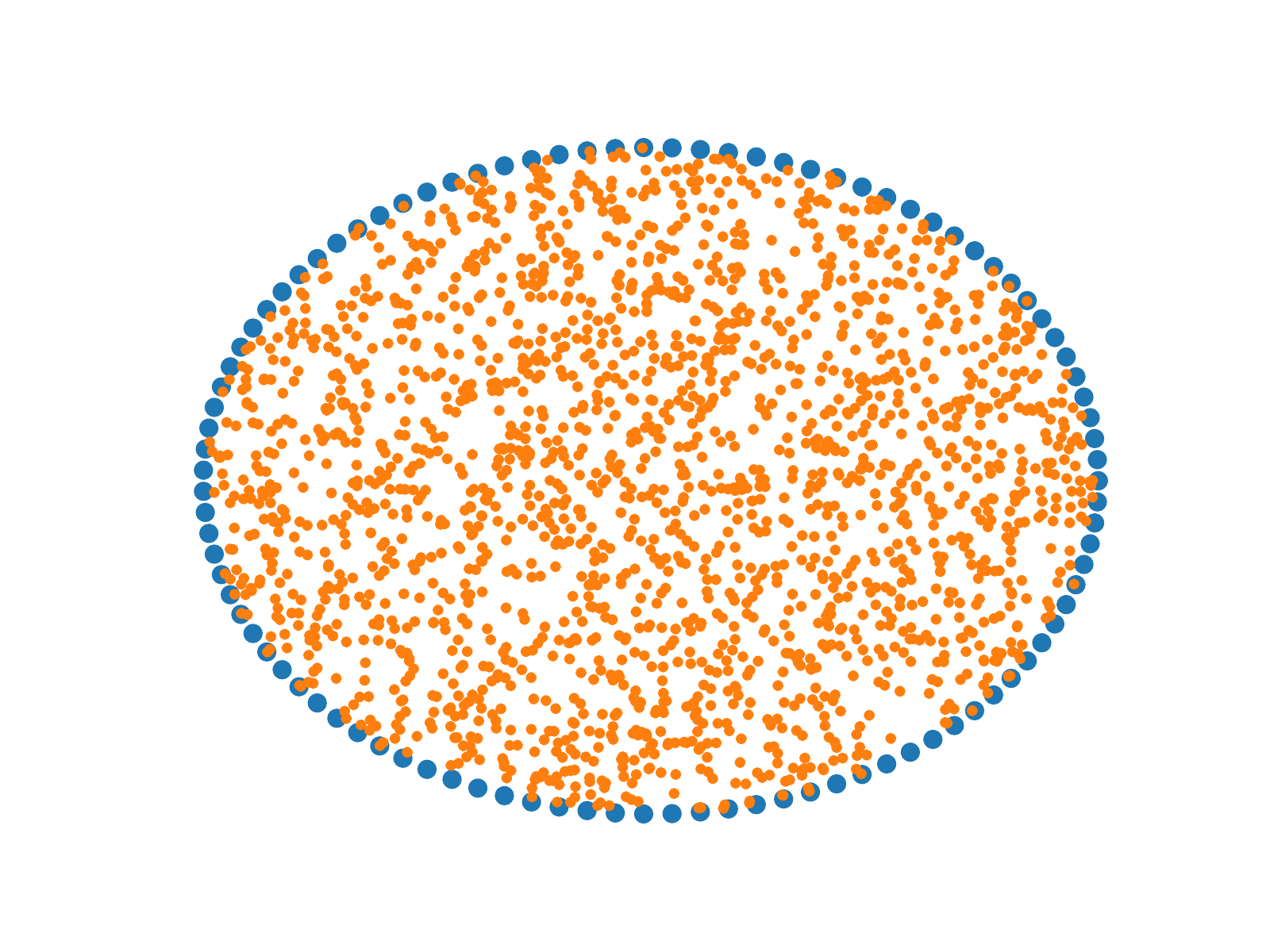}
        \caption{Sampled linearly interpolated feature representations from between only three random real feature representations at a time.}
        \label{subfig:3_lat}
    \end{subfigure}
    \caption{A visual example for demonstrating the difference in resulting diversity of sampled latent points from different sampling techniques. Sampling from a linearly interpolated combination of all real feature representations (\ref{subfig:60_lat}) results in a bias towards the centre of the feature space; whereas sampling from three randomly selected real feature representations  (\ref{subfig:3_lat}) results in a more uniform and diverse sampling.}
    \label{fig:latent_sampling}
\end{figure}

\subsubsection{Investigating Synthetic Image Diversity}
Inspired by works in domain randomisation \cite{yueDomainRandomizationPyramid2019, tobinDomainRandomizationTransferring2017} and data augmentation \cite{shortenSurveyImageData2019}, which demonstrate that increasing diversity in the training data improves performance. We investigate if diversity impacts the quality of a synthetic training dataset. In order to isolate the impact of diversity, we utilise the Real Guidance (RG) technique used in \cite{he_is_2022}, with a slight modification, to minimise the potential domain gap. RG minimises the domain gap by replacing the Gaussian noise used in the diffusion process with a real image overlayed with Gaussian noise. We modify this method by generating images from the interpolated feature representations of real images instead of from the real images themselves. More explicitly, we randomly sample 1\% of the images for each class in CIFAR10~\cite{krizhevsky_learning_2009}, totalling 60 images per class. These images were then encoded using CLIP's image encoder to obtain their feature representations. Next, we perform linear interpolation of these representations and use the interpolated feature representation as conditioning for the diffusion image generation process. Figure \ref{fig:latentDiversity} shows an example of this where we generate synthetic images of trucks (Fig. \ref{subfig:syn_images}) from the feature representations of the original truck images (Fig. \ref{subfig:orig_images}). The first row in Figure \ref{subfig:syn_images} are images generated from only the feature representation of the first image in Figure \ref{subfig:orig_images}, the second row is from the average of the first and second images and the third is average of all three. We can qualitatively observe improved synthetic image diversity from rows one to three. In row four we generate images from random interpolations between all three original images. 

We use our 60 initial feature representations to generate images via linear interpolation using two sampling methods. Firstly, we sample a starting feature representation from within the convex hull of all 60 feature representations. Secondly, we sample from only three random feature representations.  We then train a ResNet-50~\cite{he_deep_2015} model on these two generated datasets and validate performance on the real CIFAR10~\cite{krizhevsky_learning_2009} test set. The average of all latent combinations obtains \textbf{35.01\%} top-1 test accuracy and the three latent combinations obtain \textbf{52.6\%}. This shows data diversity is important when generating datasets and improving zero-shot test accuracy is dependent on generating diverse training examples. 

\newcommand{\CIFARScale}{0.13}
\begin{table*}[]
\resizebox{\textwidth}{!}{
\begin{tabular}{r@{}c@{}c@{}c@{}c@{}c@{}c@{}c@{}c@{}c@{}c@{}c@{}}
         & Airplane & Automobile & Bird    & Cat     & Deer    & Dog     & Frog    & Horse   & Ship    & Truck   \\
\multirow{2}{*}[2em]{\rotatebox[origin=c]{90}{CIFAR10~\cite{krizhevsky_learning_2009}}}  &  \includegraphics[width=\CIFARScale\textwidth]{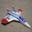}
            & \includegraphics[width=\CIFARScale\textwidth]{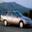}
            & \includegraphics[width=\CIFARScale\textwidth]{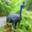}
            & \includegraphics[width=\CIFARScale\textwidth]{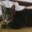}
            & \includegraphics[width=\CIFARScale\textwidth]{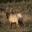}
            & \includegraphics[width=\CIFARScale\textwidth]{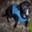}
            & \includegraphics[width=\CIFARScale\textwidth]{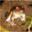}
            & \includegraphics[width=\CIFARScale\textwidth]{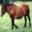}
            & \includegraphics[width=\CIFARScale\textwidth]{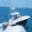}
            & \includegraphics[width=\CIFARScale\textwidth]{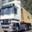} \\
         & \includegraphics[width=\CIFARScale\textwidth]{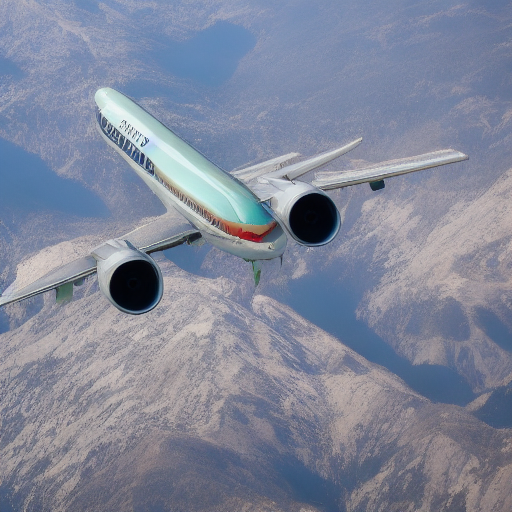}
            & \includegraphics[width=\CIFARScale\textwidth]{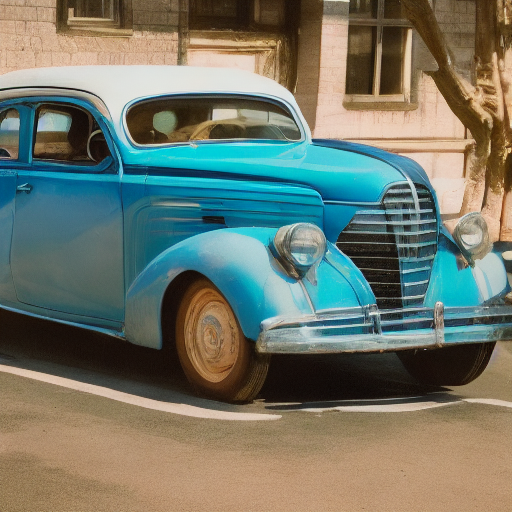}
            & \includegraphics[width=\CIFARScale\textwidth]{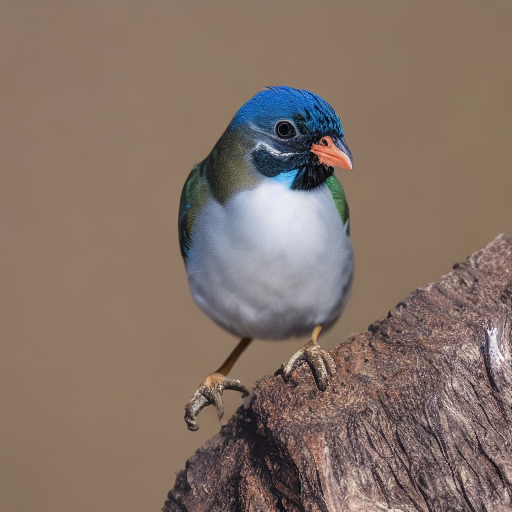}
            & \includegraphics[width=\CIFARScale\textwidth]{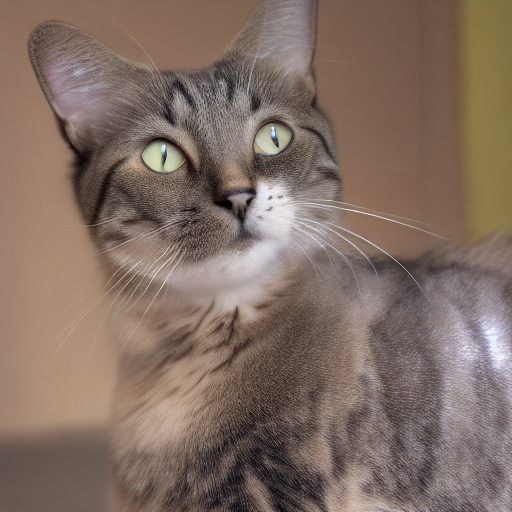}
            & \includegraphics[width=\CIFARScale\textwidth]{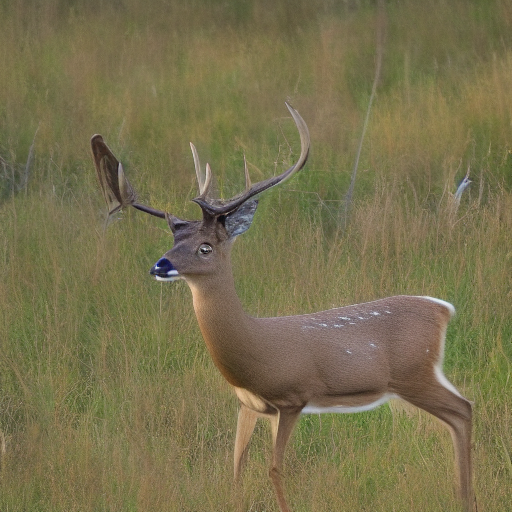}
            & \includegraphics[width=\CIFARScale\textwidth]{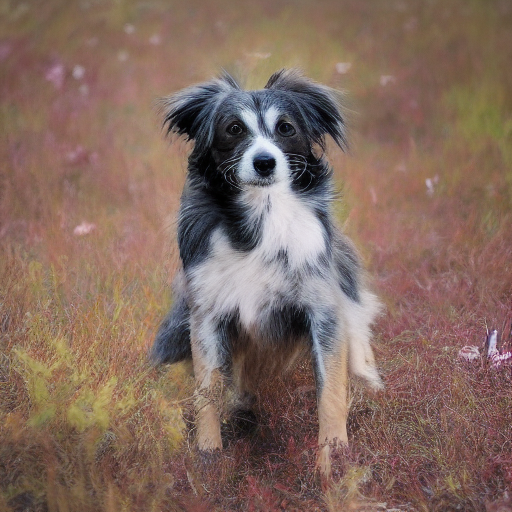}
            & \includegraphics[width=\CIFARScale\textwidth]{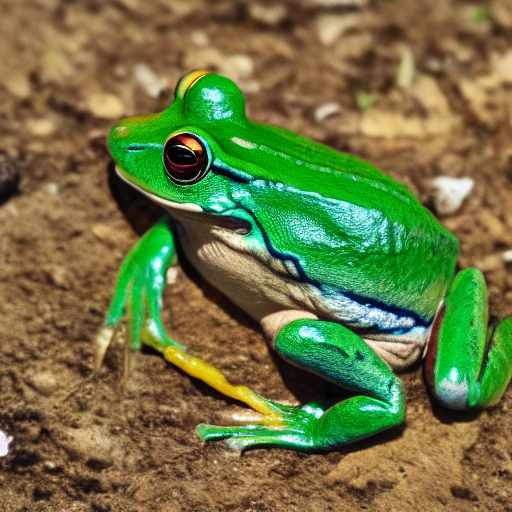}
            & \includegraphics[width=\CIFARScale\textwidth]{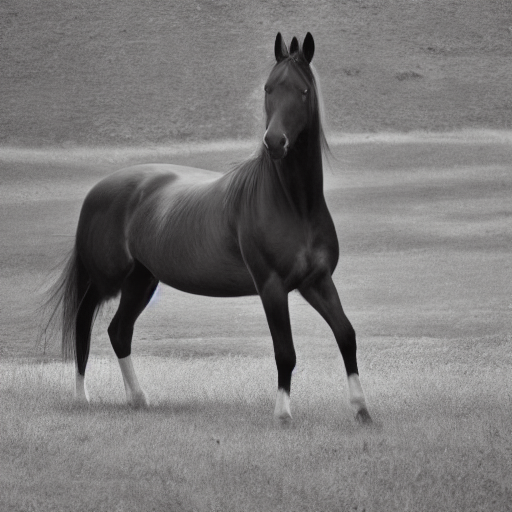}
            & \includegraphics[width=\CIFARScale\textwidth]{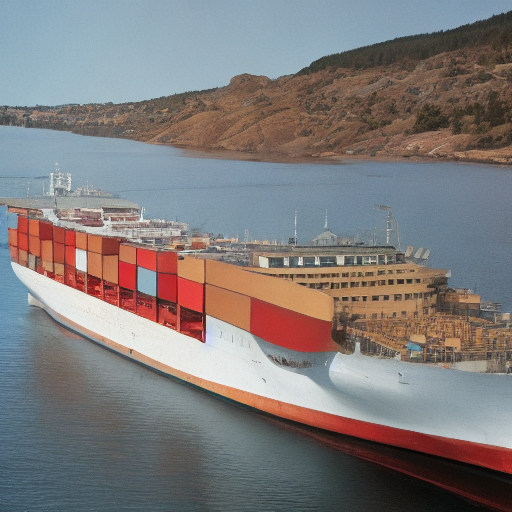}
            & \includegraphics[width=\CIFARScale\textwidth]{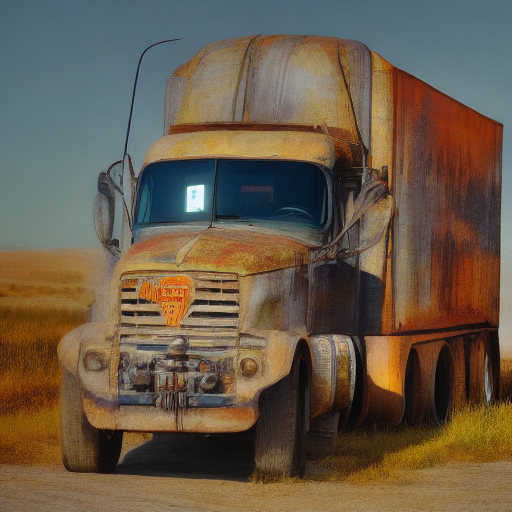}\\  \midrule
 &
  \begin{tabular}[c]{@{}c@{}}Aquarium\\ Fish\end{tabular} &
  Bicycle &
  Castle &
  Dinosaur &
  Keyboard &
  Sea &
  Shark &
  Television &
  Tractor &
  Wolf \\
\multirow{2}{*}[2em]{\rotatebox[origin=c]{90}{CIFAR100~\cite{krizhevsky_learning_2009}}} & \includegraphics[width=\CIFARScale\textwidth]{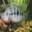}
            & \includegraphics[width=\CIFARScale\textwidth]{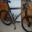}
            & \includegraphics[width=\CIFARScale\textwidth]{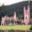}
            & \includegraphics[width=\CIFARScale\textwidth]{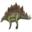}
            & \includegraphics[width=\CIFARScale\textwidth]{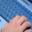}
            & \includegraphics[width=\CIFARScale\textwidth]{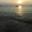}
            & \includegraphics[width=\CIFARScale\textwidth]{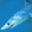}
            & \includegraphics[width=\CIFARScale\textwidth]{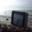}
            & \includegraphics[width=\CIFARScale\textwidth]{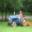}
            & \includegraphics[width=\CIFARScale\textwidth]{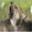} \\
         & \includegraphics[width=\CIFARScale\textwidth]{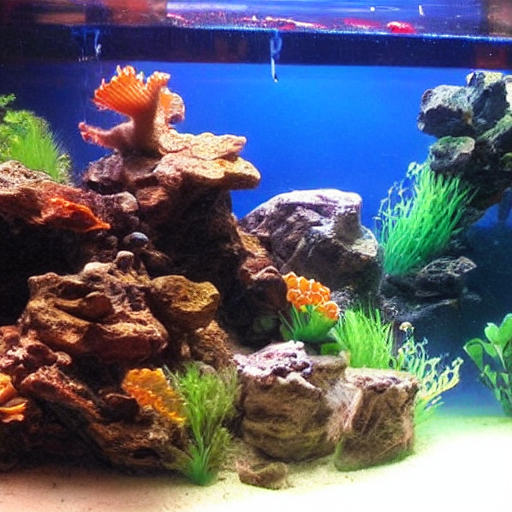}
            & \includegraphics[width=\CIFARScale\textwidth]{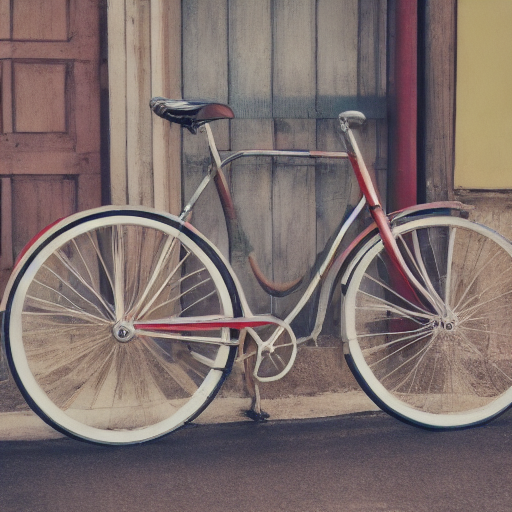}
            & \includegraphics[width=\CIFARScale\textwidth]{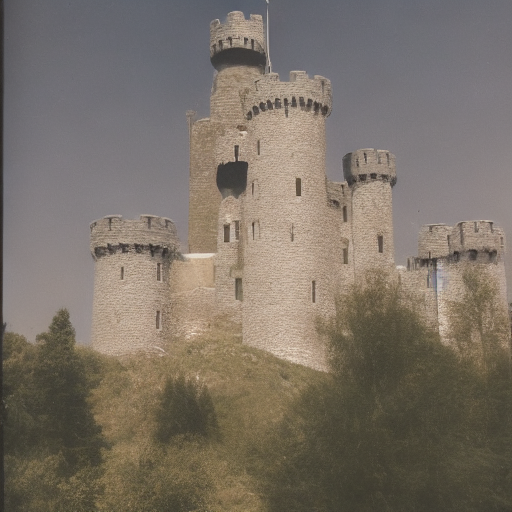}
            & \includegraphics[width=\CIFARScale\textwidth]{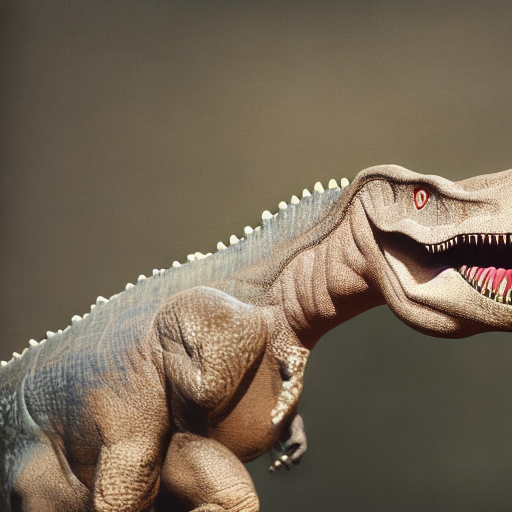}
            & \includegraphics[width=\CIFARScale\textwidth]{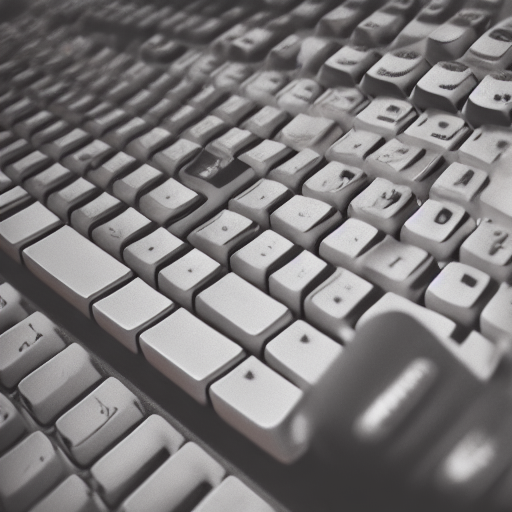}
            & \includegraphics[width=\CIFARScale\textwidth]{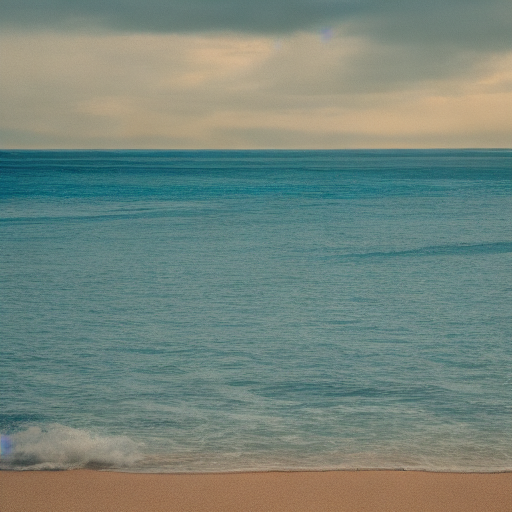}
            & \includegraphics[width=\CIFARScale\textwidth]{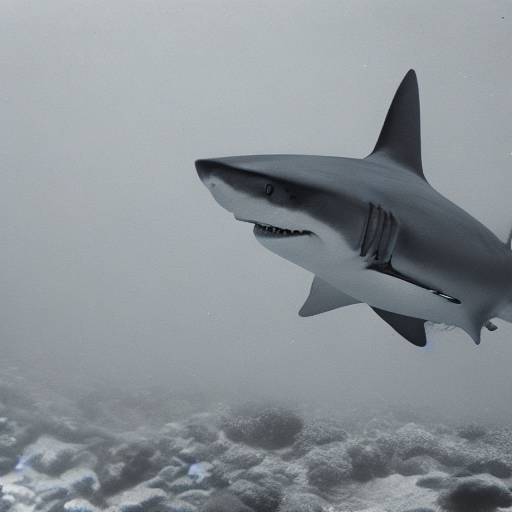}
            & \includegraphics[width=\CIFARScale\textwidth]{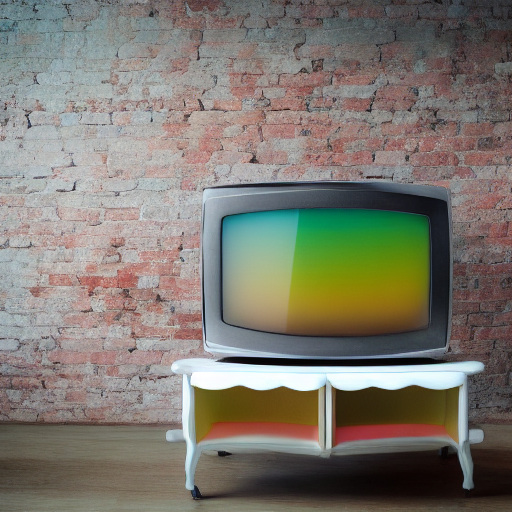}
            & \includegraphics[width=\CIFARScale\textwidth]{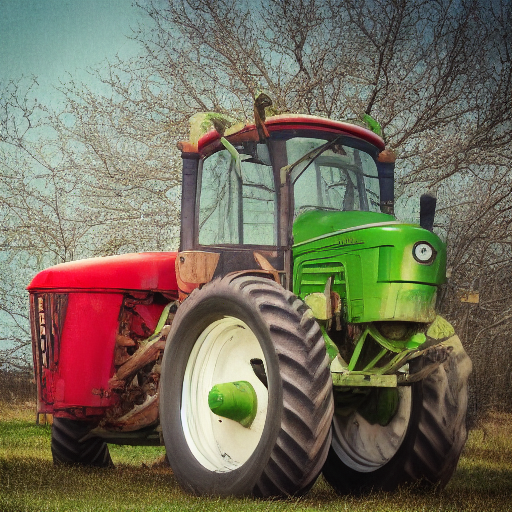}
            & \includegraphics[width=\CIFARScale\textwidth]{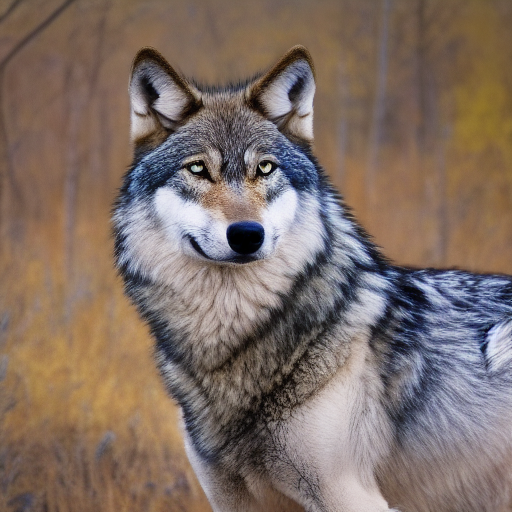} \\ \midrule
 &
  \begin{tabular}[c]{@{}c@{}}Annual\\ Crop\end{tabular} &
  Forest &
  \begin{tabular}[c]{@{}c@{}}Herbaceous\\ Vegetation\end{tabular} &
  Highway &
  Industrial &
  Pasture &
  \begin{tabular}[c]{@{}c@{}}Permanent\\ Crop\end{tabular} &
  Residential &
  River &
  Sea/Lake \\
\multirow{2}{*}[2em]{\rotatebox[origin=c]{90}{EuroSAT~\cite{helber_Eurosat_2019}}}  & \includegraphics[width=\CIFARScale\textwidth]{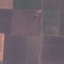}
            & \includegraphics[width=\CIFARScale\textwidth]{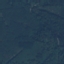}
            & \includegraphics[width=\CIFARScale\textwidth]{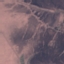}
            & \includegraphics[width=\CIFARScale\textwidth]{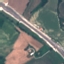}
            & \includegraphics[width=\CIFARScale\textwidth]{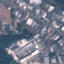}
            & \includegraphics[width=\CIFARScale\textwidth]{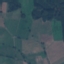}
            & \includegraphics[width=\CIFARScale\textwidth]{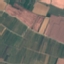}
            & \includegraphics[width=\CIFARScale\textwidth]{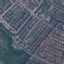}
            & \includegraphics[width=\CIFARScale\textwidth]{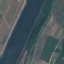}
            & \includegraphics[width=\CIFARScale\textwidth]{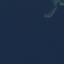} \\
         & \includegraphics[width=\CIFARScale\textwidth]{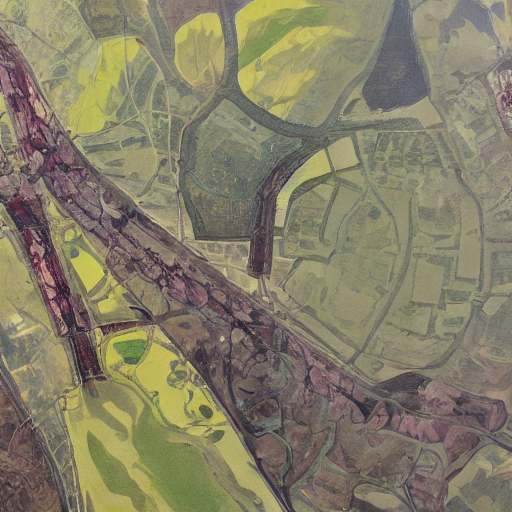}
            & \includegraphics[width=\CIFARScale\textwidth]{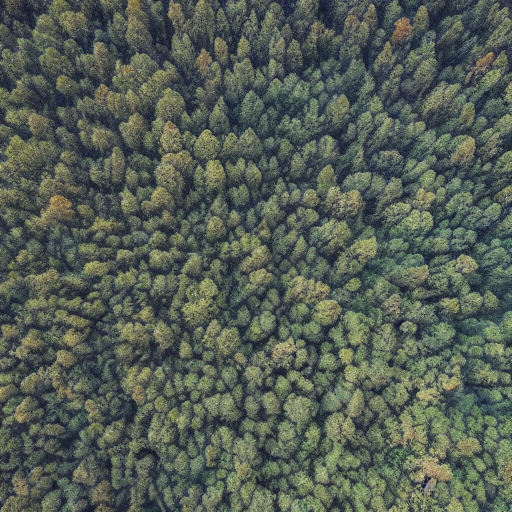}
            & \includegraphics[width=\CIFARScale\textwidth]{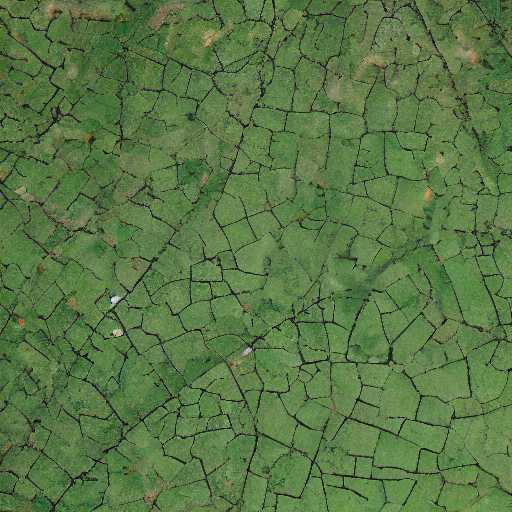}
            & \includegraphics[width=\CIFARScale\textwidth]{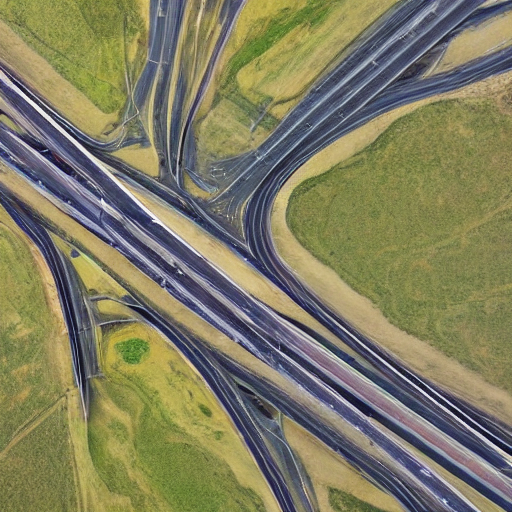}
            & \includegraphics[width=\CIFARScale\textwidth]{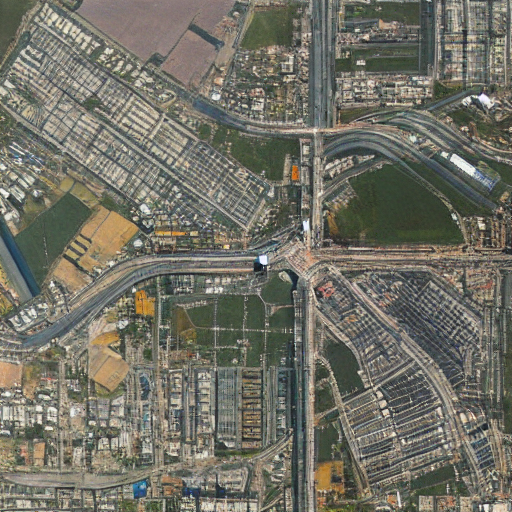}
            & \includegraphics[width=\CIFARScale\textwidth]{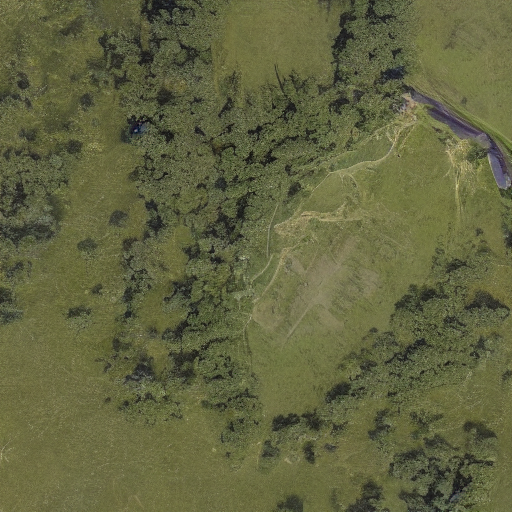}
            & \includegraphics[width=\CIFARScale\textwidth]{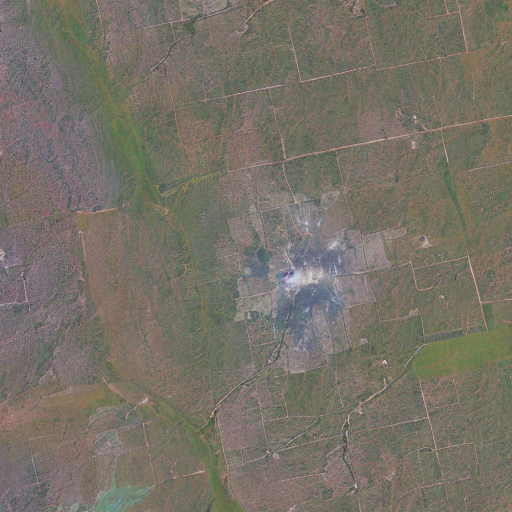}
            & \includegraphics[width=\CIFARScale\textwidth]{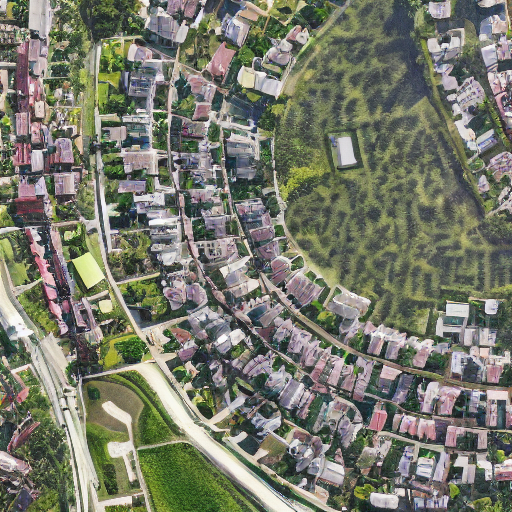}
            & \includegraphics[width=\CIFARScale\textwidth]{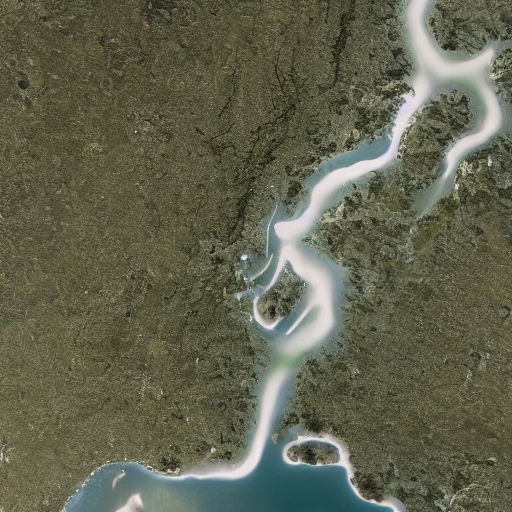}
            & \includegraphics[width=\CIFARScale\textwidth]{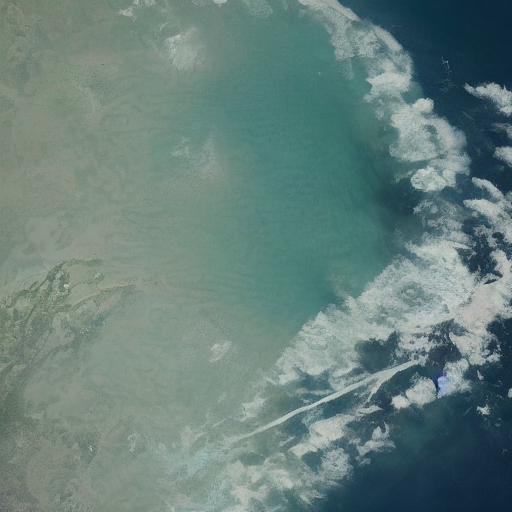}
\end{tabular}}
\caption{Examples of real (top row per dataset) and generated (bottom row per dataset) images for the CIFAR10~\cite{krizhevsky_learning_2009}, CIFAR100~\cite{krizhevsky_learning_2009} and EuroSAT~\cite{helber_Eurosat_2019}, the generated images are taken from the base class generated dataset. The real images are 32x32 for CIFAR and 64x64 for EuroSAT, while the synthetic images are 512x512. The generated images are downsized to match the real images during training.}
\label{tab:examples}
\end{table*}

In Figure~\ref{fig:latent_sampling}, we provide a visual example that more intuitively illustrates the differences between the two linear interpolation sampling methods. This is purely for demonstration purposes and does not represent the actual feature representations. We first generate a number of evenly spaced points (shown in blue) along the circumference of a circle, representing the feature representations of real images. We then use the two sampling methods described earlier to generate linearly interpolated feature representations (shown in orange). Figure~\ref{subfig:60_lat} shows a significant bias towards the centre of the simulated feature space, resulting in less diverse samples. On the other hand, Figure~\ref{subfig:3_lat} shows a more uniform and diverse set of samples.

\subsection{Bag of Tricks for Improving Synthetic Diversity}
\label{sec:bag_of_tricks}
Here we propose the tricks we can utilise to improve the diversity of generated images. We first describe these tricks below and later validate their effectiveness in improving zero-shot classification performance. For all synthetic datasets, we generate the same number of images per class as their real counterparts. 

\noindent\textbf{Base Class - }This consists of images that were generated using the prompt ``an image of a \{class\}" where \{class\} is replaced with a class name from the downstream dataset. This represents the naive case for generating images as this prompt is synonymous with the prompt used for zero-shot prediction in \cite{radford_learning_2021}. We only slightly modify the zero-shot prompt from ``a photo of" to ``an image of" as we use the prompt ``a photo of" later in the \textit{mutli-domain} trick.

\noindent\textbf{Class Prompt - }We change the prompt from ``an image of a \{class\}" to just ``\{class\}" as images generated from ``an image of a \{class\}" are included in the subset of images generated by the prompt ``\{class\}". Therefore, using just the class name may lead to more diverse outputs.

\noindent\textbf{Multi-Domain - }Next, we directly influence the diversity by providing a list of domains with the prompt ``a \{domain\} of a \{class\}" where the domain is one of ten preset domains (photo,  drawing, painting, sketch, collage, poster, digital art image, rock drawing, stick figure, 3D rendering) for CIFAR datasets. Due to EuroSAT requiring more domain information to correctly generate a satellite image, we use the prompt ``a satellite photo of a \{class\} in the style of a \{domain\}" where the domains are (realistic photo, drawing, painting, sketch, 3D rendering). The images in Figure \ref{fig:txt2img} are an example of CIFAR10 \textit{multi-domain} images.

\noindent\textbf{Random Unconditional Guidance - }We use the \textit{base class} prompt and randomly set the unconditional guidance scale between values of 1 and 5. This generates images that are highly diverse, such as generated with $UCG = 1$, as well as images containing stronger features of the target class, such as with $UCG = 5$. $UCG = 5$ was chosen as the upper bound from qualitative inspection of generated images where we found little difference in features of synthetic images with values greater than 5.

\noindent\textbf{All Combined - }Lastly, we combine all previous tricks into one final dataset. This should result in a more diverse dataset than any individual dataset.

\section{EXPERIMENTS}
\label{sec:experiments}
In this section, we first discuss our training setup and then present our baseline zero-shot results with ReseNet50 \cite{he_deep_2015}. We then apply the \textit{bag of tricks} when generating our synthic datasets of CIFAR10 \cite{krizhevsky_learning_2009}, CIFAR100 \cite{krizhevsky_learning_2009} and EuroSAT \cite{helber_Eurosat_2019}. We have chosen the CIFAR datasets due to their widespread use and low resolution, allowing for ease of training. We have also included EuroSAT because of its low resolution and challenging domain for synthetic images. Additionally, it has been shown to be a challenging dataset for zero-shot classification~\cite {radford_learning_2021}. We show examples of our generated datasets in Table \ref{tab:examples}.

Following the \textit{bag of tricks}, we have identified the the best tricks for each dataset and tested them on four additional classification architectures; ResNet101 \cite{he_deep_2015}, MobileNetV3 \cite{howard_searching_2019}, ViT \cite{dosovitskiy_image_2021} and ConvNeXt \cite{liu_convnet_2022}.

\subsection{Training Setup}
All models were trained from random initialisation for 200 epochs, with a batch size of 128, the AdamW optimiser \cite{loshchilov2018decoupled} and cosine annealing learning rate decay. All training used an initial learning rate of $2e^{-4}$. MobileNetV3 models used a weight decay of 0.1 for all training; whereas all other models used a weight decay of 0.9 when training on the CIFAR datasets and 0.3 when training on EuroSAT. We found the weight decay hyperparameter of the AdamW optimiser to be important, as it helps reduce overfitting to the synthetic images.

\begin{table}[t]
\centering

\resizebox{\columnwidth}{!}{%
\begin{tabular}{llll}
\hline
                               & CIFAR10 \cite{krizhevsky_learning_2009} & CIFAR100 \cite{krizhevsky_learning_2009}  & EuroSAT \cite{helber_Eurosat_2019} \\ \hline
\textbf{He et al. (ResNet50)} \cite{he_is_2022}             & -       & 28.74       & -   \\ 
\textbf{CLIP-ResNet50} \cite{radford_learning_2021}             & 75.6    & 41.6 & 41.1   \\ 
\textbf{Base Class}            & 60.5    & 29.72                  & 36.18   \\\hline
\textbf{Anti-aliasing Rescale} & 63.84  \textcolor{Green}{\(\uparrow\)3.34}  & 33.61 \textcolor{Green}{\(\uparrow\)3.89} & 34.4  \textcolor{Red}{\(\downarrow\)1.78}  \\
\textbf{Class Prompt}    & 62.32  \textcolor{Green}{\(\uparrow\)1.82}  & 26.4 \textcolor{Red}{\(\downarrow\)3.32}   & -     \\
\textbf{Multi-Domain}    & 67.97  \textcolor{Green}{\(\uparrow\)7.47}  & 32.55  \textcolor{Green}{\(\uparrow\)1.96} & 35.68  \textcolor{Red}{\(\downarrow\)0.5}   \\
\textbf{Random Guidance} & 72.93  \textcolor{Green}{\(\uparrow\)12.43} & 31.19  \textcolor{Green}{\(\uparrow\)1.47} & \textbf{40.18} \textcolor{Green}{\(\uparrow\)4} \\
\textbf{All Combined}    & \textbf{81}  \textcolor{Green}{\(\uparrow\)20.5} & \textbf{45.63}  \textcolor{Green}{\(\uparrow\)15.91} & 39.92 \textcolor{Green}{\(\uparrow\)3.74}\\ \hline
\end{tabular}%
}

\caption{Zero-shot classification top-1 test accuracy on the CIFAR10, CIFAR100 and EuroSAT datasets from training on different permutations of synthetic datasets with a ResNet50 model. The change (\textcolor{Red}{\(\downarrow\)}, \textcolor{Green}{\(\uparrow\)}) in top-1 accuracy is measured with respect to the \textit{base class}.}
\label{tab:tricks_res}
\end{table}

\subsection{Baseline Results}
In Table \ref{tab:tricks_res}, we gather baseline zero-shot classification results of CLIP-ResNet50 from \cite{radford_learning_2021} and ResNet50 from \cite{he_is_2022}. For our own baseline, we use the \textit{base class} synthetic dataset as described in section \ref{sec:bag_of_tricks}. Our \textit{base class} dataset achieves zero-shot accuracy of 60.5\%, 29.72\% and 36.18\% for CIFAR10, CIFAR100 and EuroSAT respectively. In comparison, CLIP-ResNet50 from \cite{radford_learning_2021} achieves zero-shot accuracies of 75.6\%, 41.6\% and 41.1\% respectively. The 15.1\% and 11.88\% difference between CLIP's ResNet zero-shot and our ResNet zero-shot results show that our generated dataset does not currently capture the full diversity of CLIP's knowledge of each class. In theory, we expect that with infinite training examples, we should achieve CLIP's zero-shot accuracy.  The baseline results from \cite{he_is_2022} on CIFAR100 are the most directly comparable to our \textit{base class} results as these results are obtained from training a ResNet50 model from scratch on a synthetic version of CIFAR100 generated using GLIDE \cite{he_is_2022}. An important point to note however is that the results in \cite{he_is_2022} are obtained after improving the prompt quality for generating synthetic datasets, whereas our \textit{base class} results are already higher without any improvements. This is most likely due to the difference in diffusion models used for the generation of the synthetic datasets. As mentioned, \cite{he_is_2022} use GLIDE \cite{nichol_glide_2022} which was trained on a dataset of 250 million image and caption pairs. In contrast, we use Stable Diffusion, which was trained on 2.3 billion image caption pairs from the LAION-5B dataset \cite{schuhmann_laion-5b_nodate}. This 9.2$\times$ increase in training data appears to result in inherently better generative abilities.

\subsection{Implementing the Bag of Tricks}
Here we iterate over the \textit{bag of tricks}, as described in Section \ref{sec:bag_of_tricks}, in an effort to improve diversity and zero-shot classification. Results are shown in Table \ref{tab:tricks_res} comparing the accuracy obtained using each \textit{trick}, with a ResNet50 model. Although not a trick, we test the impact of using anti-aliasing during the rescaling of images from 512$\times$512 pixels to 32$\times$32 for CIFAR datasets and 64$\times$64 for EuroSAT. We find anti-aliasing significantly benefits CIFAR datasets but not EuroSAT. Thus we do not apply anti-aliasing on EuroSAT.

\noindent\textbf{Class Prompt - }Using only the class name as the prompt we see a small improvement in CIFAR10 accuracy while CIFAR100 accuracy reduces. We suspect the reduction in CIFAR100 performance is due to the generation of incorrect images for classes which can have multiple meanings. Such as:
`Apple' generating images of the fruit and the software company Apple logo.
`Beetle' generating images of the insect and the Volkswagen car.
`Orange' generating images of orange items of clothing mainly instead of the fruit.
`Ray' generating Sun rays, sting rays and men.
This is not the case for CIFAR10 where the class names are unambiguous. 
We do not test this trick on the EuroSAT dataset as this dataset requires some context in the prompt relating to satellite images.

\noindent\textbf{Multi-Domain - }Despite approximately 90\% of the images being generated under this setting not being realistic; we see the most significant improvement for the CIFAR datasets, with a 7.47\% and 1.96\% improvement with CIFAR10 and CIFAR100 respectively. These images, especially the posters, paintings and drawings, are not within the real CIFAR domain. For EuroSAT, we see a slight reduction of 0.5\% compared to the \textit{base class}. Both the CIFAR and EuroSAT results show that out-of-domain training images are not the main constraint for improving the zero-shot potential of synthetic datasets.

\noindent\textbf{Random Unconditional Guidance - }When we directly enforce diversity over precision by setting a random unconditional guidance scale we see an improvement in zero-shot classification across all datasets. Interestingly, we see the most significant improvements in CIFAR10 and EuroSAT. We conjecture this is due to CIFAR10 and EuroSAT containing more training examples per class than CIFAR100. This supports the finding in \cite{he_is_2022} that synthetic images are less data efficient than real images. Random unconditional guidance resulting in an accuracy improvement further supports our hypothesis that increasing diversity is more important than reducing a domain gap when generating synthetic training images.

\noindent\textbf{All Combined - }Finally, we combine all previously generated datasets into one large dataset, in order to test if combining all the tricks, and further increasing diversity, gives more zero-shot classification improvements. In doing so we obtain our most significant improvements, further supporting our hypothesis. Surprisingly, both CIFAR zero-shot results have now surpassed the CLIP-ResNet50 zero-shot results \cite{radford_learning_2021}, showing that our \textit{bag of tricks} may distil the important signals or features in CLIPs understanding of a concept.

\begin{table}[t]
\centering
\resizebox{\columnwidth}{!}{%
\begin{tabular}{llll}
\hline
                          &                & \textbf{Base Class} & \textbf{Best Tricks} \\ \hline
\textbf{Dataset}          & \textbf{Model} &            &             \\ \hline
\multirow{4}{*}{CIFAR10 \cite{krizhevsky_learning_2009}}   
& CLIP-ResNet50$^*$ \cite{radford_learning_2021} & 75.6 & - \\
& ResNet50 \cite{he_deep_2015}          & 60.5       & 81 \textcolor{Green}{\(\uparrow\)20.5}       \\
                          
                          & ResNet101 \cite{he_deep_2015}         & 60.89      & \textbf{81.84} \textcolor{Green}{\(\uparrow\)20.95}         \\
                          & ViT-B \cite{dosovitskiy_image_2021}             & 42.34      & 75.72 \textcolor{Green}{\(\uparrow\)33.38}      \\ 
                          & MobileNetV3-S \cite{howard_searching_2019}       & 51.05      & 74.38 \textcolor{Green}{\(\uparrow\)23.33}      \\
                          & ConvNeXt-S \cite{liu_convnet_2022} & 49.15  & 80.1 \textcolor{Green}{\(\uparrow\)30.95} \\\hline
\multirow{4}{*}{CIFAR100 \cite{krizhevsky_learning_2009}} 
& CLIP-ResNet50$^*$ \cite{radford_learning_2021} & 41.6 & - \\
& ResNet50$^*$ \cite{he_is_2022} & 28.74 & - \\
&ResNet50          & 29.72      & 45.63 \textcolor{Green}{\(\uparrow\)15.91}       \\
                          & ResNet101          & 27.66      & \textbf{46.63}  \textcolor{Green}{\(\uparrow\)18.97}     \\
                          & ViT-B              & 16.38      & 32.38  \textcolor{Green}{\(\uparrow\)16}     \\ 
                          & MobileNetV3-S        & 17.78      & 39.64 \textcolor{Green}{\(\uparrow\)21.86}      \\
                          & ConvNeXt-S & 20.93 & 45.14 \textcolor{Green}{\(\uparrow\)24.21}\\\hline
\multirow{4}{*}{EuroSAT \cite{helber_Eurosat_2019}}  
& CLIP-ResNet50$^*$ \cite{radford_learning_2021} & 41.1 & - \\
& ResNet50           & 36.18      & \textbf{42.59} \textcolor{Green}{\(\uparrow\)6.41}      \\
                          & ResNet101          & 34.73      & 37.31 \textcolor{Green}{\(\uparrow\)2.58}      \\
                          & ViT-B              & 19.53      & 21.71 \textcolor{Green}{\(\uparrow\)2.18}      \\
                          & MobileNetV3-S        & 34.08      & 39.13 \textcolor{Green}{\(\uparrow\)5.05}      \\
                          & ConvNeXt-S & 18.57 & 20.22 \textcolor{Green}{\(\uparrow\)1.65}
\end{tabular}
}

\caption{Top-1 zero-shot accuracy of various classification models on the \textit{Base Class} and \textit{Best Trick} synthetic CIFAR10, CIFAR100 and EuroSAT datasets. Models marked with $^*$ are baseline results (not \textit{base class} results) from the cited papers, similar to Table \ref{tab:tricks_res}. The change (\textcolor{Green}{\(\uparrow\)}) in \textit{Best Trick} top-1 accuracy is relative to the \textit{Base Class} top-1 accuracy.}
\label{tab:best_tricks}
\end{table}

\subsection{Model Agnostic Zero-shot Classification}
Using our \textit{bag of tricks} we can now endow any model with zero-shot classification capabilities. To demonstrate this we test the best tricks for each dataset on four additional classification architectures, results are shown in Table \ref{tab:best_tricks}. For CIFAR10, the \textit{all combined} dataset is used for the \textit{best tricks}. CIFAR100 uses all tricks except \textit{class prompt}. For EuroSAT, only the \textit{random unconditional guidance} trick improved performance, therefore to further increase diversity, we generate an additional 2700 images per class, doubling the size of the dataset.  We use the ResNet101 \cite{he_deep_2015} architecture in order to test if simply a deeper ResNet is able to obtain higher zero-shot performance and we see only slight improvements over ResNet50. When training with the ViT-B model \cite{dosovitskiy_image_2021} we see reduced performance compared to the ResNet models across all datasets. We conjecture this is due to training from scratch, as ViTs are known to benefit greatly from ImageNet pre-training \cite{dosovitskiy_image_2021}. Despite this, we still see an improvement in zero-shot performance when training using the \textit{best tricks}. Lastly, we use MobileNetV3-small \cite{howard_searching_2019} and ConvNeXt-small \cite{liu_convnet_2022} as examples of architectures that previously have not been used for zero-shot classification, demonstrating our approach applies to any existing model. Again we see improvements in zero-shot classification across all datasets when applying the \textit{best tricks}.

\section{Conclusion}
In conclusion, we investigate the problem of Model-Agnostic Zero-Shot Classification (MA-ZSC). Where MA-ZSC aims to train any downstream classification architecture to classify real images without training on any real images. We investigated how to improve the quality of a synthetic dataset for the purpose of training and found diversity in the synthetic images to be an important factor. From this, we then proposed a set of modifications to the text-to-image generation process via diffusion models, named our \textit{bag of tricks}. This \textit{bag of tricks} is designed only to improve the diversity of synthetic images, with no mitigation of the potential domain gap, as reported by previous works. Applying the \textit{bag of tricks} achieves notable improvements across five classification architectures on the CIFAR10, CIFAR100 and EuroSAT datasets. Some architectures even achieve zero-shot classification accuracies comparable to state-of-the-art zero-shot models, such as CLIP. Our findings provide initial insights into the problem of MA-ZSC using diffusion models and opens up new avenues for research in this area.

\vspace{-3mm}\section*{Acknowledgement}\vspace{-2mm}
This work has been supported by the SmartSat CRC, whose activities are funded by the Australian Government’s CRC Program; and partly supported by Sentient Vision Systems. Sentient Vision Systems is one of the leading Australian developers of computer vision and artificial intelligence software solutions for defence and civilian applications.

{\small
\bibliographystyle{ieee_fullname}
\bibliography{references}
}

\end{document}